\documentclass{article}

\usepackage{arxiv}

\usepackage[utf8]{inputenc} 
\usepackage[T1]{fontenc}    
\usepackage{hyperref}       
\usepackage{url}            
\usepackage{booktabs}       
\usepackage{amsfonts}       
\usepackage{nicefrac}       
\usepackage{microtype}      
\usepackage{lipsum}		
\usepackage{graphicx}
\usepackage{natbib}
\usepackage{doi}
\usepackage{subcaption}

\title{Diversidade linguística e inclusão digital: \\ desafios para uma IA brasileira}

\author{ \href{https://orcid.org/0000-0002-4972-4320}{\includegraphics[scale=0.06]{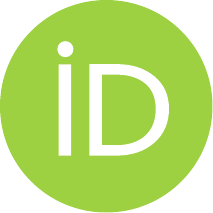}\hspace{1mm}Raquel Meister Ko. Freitag}\\
	Departamento de Letras Vernáculas \\ Universidade Federal de Sergipe -  UFS\\
	\texttt{rkofreitag@academico.ufs.br} \\
}

\date{}

\renewcommand{\headeright}{}
\renewcommand{\shorttitle}{Diversidade linguística e inclusão digital: \\ desafios para uma IA brasileira}

\begin{document}
\maketitle

\begin{abstract}
    Linguistic diversity is a human attribute which, with the advance of generative AIs, is coming under threat. This paper, based on the contributions of sociolinguistics, examines the consequences of the variety selection bias imposed by technological applications and the vicious circle of preserving a variety that becomes dominant and standardized because it has linguistic documentation to feed the large language models for machine learning. 
\end{abstract}


\section{Introdução}

No documento do Ministério da Ciência, Tecnologia e Inovação \textit{IA para o Bem de Todos}, em que é apresentada a proposta de um 
\textit{Plano Brasileiro de Inteligência Artificial 2024-2028}, um dos cinco objetivos listados é “Desenvolver modelos avançados de linguagem em português, com dados nacionais que abarcam nossa diversidade cultural, social e linguística, para fortalecer a soberania em IA.” \cite{IAbem:2024}. 

A Sociolinguística é o campo da ciência que estuda as relações entre língua e sociedade, e o conjunto de trabalhos neste campo desenvolvidos no Brasil nos últimos 50 anos tem contribuições diretas para a consecução deste objetivo \cite{freitag2016sociolinguistica}. E é sob esta perspectiva que este objetivo é discutido neste texto. Pensando em uma IA ética e socialmente sensível, a diversidade das comunidades na sociedade se reflete também (ou, pelo menos, deveria se refletir) na diversidade das comunidades em amostras linguísticas para treinar modelos de língua em larga escala (LLMS).

Uma IA ética precisa atender aos princípios de justiça, equidade, diversidade e inclusão, e no domínio linguístico, por meio da seleção das amostras de línguas e variedades de línguas que vão compor o corpus de treno dos modelos, assimetrias se acentuam, desde a exclusão ou apagamento de línguas, até a priorização de uma variedade -- dita de prestígio -- face às variedades consideradas não-padrão ou estigmatizadas. Os preconceitos decorrentes dessa hieraquização de variedades são reproduzidos em LLMs e geram respostas \cite{shrawgi2024uncovering, fleisig2024linguistic, freitag2024performance}, como já constatado no inglês afro-americano \cite{mengesha2021don, dacon2021truly, dacon2022evaluating}.  

Considerando o objetivo do \textit{Plano Brasileiro de Inteligência Artificial 2024-2028} que trata de diversidade linguística, primeiramente, o mito do monolinguismo do português precisa ser desfeito. Em seguida, o português falado no Brasil é apresentado sob a perspectiva da diversidade e a tensão entre variedades de prestígio e variedades ditas "não-padrão" que divide a sociedade. Após essa contextualização sociolinguística, são apresentadas recomendações para a constituição de amostras linguísticas brasileiras para treinar LLMs, de modo a garantir a diversidade cultural, social e linguística prevista na proposta do \textit{Plano Brasileiro de Inteligência Artificial}.

\section{No Brasil, não se fala só português}

Sobre língua, a Constituição de 1988 reconhece, no Art. 13., que "A língua portuguesa é o idioma oficial da República Federativa do Brasil." \cite{const}. O objetivo do \textit{Plano Brasileiro de Inteligência Artificial 2024-2028} reflete o dispositivo legal. No entanto, não é apenas português que se fala no Brasil. A existência de outras línguas, embora empírica e legalmente reconhecidas, não faz parte do imaginário da nação, que se molda por uma ideologia monolíngue -- a de que aqui todos falamos português -- que se reproduz nos LLMs, na medida que somente o português é reconhecido como língua de soberania nacional no documento norterador. 

Na própria constituição, bem mais distante, há pistas da diversidade linguística, como no § 2º do Art. 210, que garante que "O ensino fundamental regular será ministrado em língua portuguesa, assegurada às comunidades indígenas também a utilização de suas línguas maternas e processos próprios de aprendizagem.", ou, ainda mais longe, no Art. 231., que diz que "São reconhecidos aos índios sua organização social, costumes, línguas, crenças e tradições, e os direitos originários sobre as terras que tradicionalmente ocupam, competindo à União demarcá-las, proteger e fazer respeitar todos os seus bens." Mesmo status de reconhecimento tem a Libras. O art. 1º da Lei 10.436/2002 diz que “É reconhecida como meio legal de comunicação e expressão a Língua Brasileira de Sinais - Libras e outros recursos de expressão a ela associados.” \cite{Lei:10.436:2002}. 

A co-oficialização é outro processo que reconhece legalmente as línguas. As primeiras línguas co-oficializadas foram três línguas indígenas faladas no município de São Gabriel da Cachoeira, estado do Amapá: Tukano, Baniwa e Nheengatu. Desde então, já são 23 línguas cooficializadas no país, sendo 13 línguas indígenas e 9 e imigração \cite{freitag2023contatos}.

E, pela vertente da patrimonialização, reconhecimento e valorização da diversidade linguística brasileira, o Instituto do Patrimônio Histórico e Artístico Nacional (Iphan), por meio do Inventário Nacional da Diversidade Linguística (INDL), tem atuado na “identificação, documentação, reconhecimento e valorização das línguas portadoras de referência à identidade, à ação e à memória dos diferentes grupos formadores da sociedade brasileira” \cite{Decreto:7.387:2010}. As línguas do Brasil, no escopo do INDL, são de seis grupos: indígenas, comunidades afro-brasileiras, imigração, sinais, crioulas e a Língua Portuguesa e suas variações dialetais. Já foram reconhecidas como \textit{Referência Cultural} cinco línguas de base indígena (duas línguas do tronco Tupi, Asurini e Guarani M'bya, três línguas da família Karib (Nahukuá, Matipu e Kuikuro Kalapalo), duas línguas de contato (Talian e Portunhol) e uma língua geral Nheengatu \cite{freitag2023contatos}.

Além da informação de base legal sobre a existência de línguas, estudos linguísticos identificam e documentam outras tantas, de modo que não há consenso sobre quantas línguas são faladas no Brasil, nem quantas pessoas falam cada uma dessas línguas. Há, no entanto, consenso de que no Brasil não se fala apenas português, e uma política para a soberania nacional não deve ignorar a diversidade linguística, sob pena não só de excluir os povos originários, como também de excluir a identidade de uma população socialmente diversa. 

Modelos de língua em larga escala para uma IA de soberania nacional precisam considerar a diversidade de línguas do Brasil, e não apenas eleger o português como língua de treino. E, mesmo dentro do português, há diversidade que reflete padrões sociais e culturais da realidade brasileira, que, como veremos na sequência, precisam ser considerados. 

\section{O português falado no Brasil é diverso}

Seja como uma das línguas com o maior número de falantes ou como uma língua com o maior número de países onde é falado, o português aparece nos ranqueamentos de línguas do mundo. O português não é apenas falado em Portugal e no Brasil \cite{meister2022sociolinguistic}. Não há um Português, há variedades de português, e cada uma destas variedades é polarizada em um centro, o que o configura o português como uma língua pluricêntrica. 

O pluricentrismo do português é reconhecido nas ações de inclusão digital: é frequente encontrar documentação de \textit{software} nas duas variedades hegemônicas do português (Português Europeu e Português Brasileiro) \cite{de2021desafios}. E, mesmo no Brasil, as especificadades de cada uma das comunidades que têm o Português como sua língua refletem seus valores socioculturais e diferenciam as variedades, o que tem sido amplamente demostrado pela sociolinguística brasileira \cite{roncarati2003portugues, abraccado2015mapeamento}. 

A diversidade do português brasileiro é reconhecida no INDL -- Língua Portuguesa e suas variações dialetais -- e também é alçada a direito de aprendizagem na Base Nacional Comum Curricular \cite{BNCC}: “Compreender as línguas como fenômeno (geo)político, histórico, cultural, social, variável, heterogêneo e sensível aos contextos de uso, reconhecendo suas variedades e vivenciando-as como formas de expressões identitárias, pessoais e coletivas, bem como agindo no enfrentamento de preconceitos de qualquer natureza”.

Assim, para a soberania nacional, uma IA brasileira não pode se limitar a uma única língua, o português, nem a uma única variedade do português. O viés de seleção de uma única língua/variedade reforça e acentua ainda mais os preconceitos, em especial contra às variedades linguísticas subrepresentadas.

\section{Recomendações para o desenvolvimento de uma IA brasileira linguisticamente diversificada}

Para cumprir o objetivo do \textit{Plano Brasileiro de Inteligência Artificial 2024-2028}, é necessário não só a intensificação de ações de documentação linguística de variedades subrepresentadas, mas a conscientização dos desenvolvedores de que as aplicações das tecnologias precisam refletir valores linguísticos e a representatividade para o grupo, sob pena de reforçar ainda mais o preconceito que já existe em relação às variedades linguísticas subrepresentadas. 

Nos seus 50 anos de trajetória, além das descrições linguísticas que caracterizam cientificamente a diferença entre a variedade brasileira e europeia do português, os estudos sociolinguísticos brasileiros vêm acumulando como produto primário um expressivo acerco de documentação linguística \cite{freitag2012bancos, freitag2021desafios, freitag2021desafios, machado2021collections, machado2021mapeamento, sousa2024bancos}. Resultado de pesquisa de campo para subsidiar teses e dissertações, estas ações de documentação linguística resultam em produtos, e como tais, têm custo e valor. Dados linguísticos autênticos, especialmente dados transcritos e anotados, têm aplicação em diversas áreas das tecnologias de linguagem e inteligência artificial. Atualmente estes acervos são armazenados de maneira assistemática e provisória, sem protocololos específicos para compartilhamento e reuso. Embora as instituições de pesquisa tenham repositórios para compartilhamento de produção científica (teses, dissertações, etc.), estes não são apropriados para compartilhar coleções de dados linguísticos. A solução para este problema é a construção de um repositório próprio específico para este tipo de acervo, em uma iniciativa denominada \textbf{Plataforma da Diversidade Linguística Brasileira} \cite{machado2021plataforma}. 

Com a articulação da Comissão de Sociolinguística da Associação Brasileira de Linguística (ABRALIN) e do GT de Sociolinguística da Associação Nacional de Pós-Graduação em Letras e Linguística (ANPOLL), a \textbf{Plataforma da Diversidade Linguística Brasileira} configura-se como um projeto nacional alinhado à visão estratégica para o desenvolvimento sustentável da área de IA, como destadado no objetivo da proposta do \textit{Plano Brasileiro de Inteligência Artificial 2024-2028}, e que demanda financiamento e institucionalização de um repositório nacional que oportunizem à sociedade a diversidade do patrimônio linguístico que vem sendo registrado e mapeado. A plataforma visa a catalogação nacional (salvaguarda e difusão)  e a constituição de um repositório comum, com padrões de metadados e diretrizes de armazenamento de de coleções de dados sociolinguísticos \cite{sousa2024bancos}, de modo a atender necessidades tanto do público amplo, para que possibilite a qualquer interessado ver, ouvir, repetir diferentes manifestações de uso linguístico no país, como para público especializado, tal como a alimentação de LLMs para uma IA brasileira.

Uma IA eticamente sensível para a soberania nacional requer que a diversidade linguística seja considerada de maneira plena equinâme, com amostras linguísticas diversificadas para o treino de LLMs. Sem isso, a reprodução de uma IA que considera apenas o português e uma de suas variedades, tem efeito na conformação de padrões linguísticos hegemônicos, invisibilizando e marginalizando ainda mais as variedades linguísticas subrepresentadas.

\bibliographystyle{unsrtnat}
\bibliography{references}

\end{document}